**Author Information:**

1. **Ramna Maqsood**

   **Email: ramnamaqsood@cuilahore.edu.pk**

   **ORCID ID: 0000-0001-9551-8926**

   **COMSATS University Islamabad Lahore Campus, Lahore Pakistan**

   **Department of Computer Science**

2. **Usama Ijaz Bajwa (Corresponding Author)**

   **Email: usamabajwa@cuilahore.edu.pk**

   **ORCID ID: 0000-0001-5755-1194**

   **Address: Department of Computer Science, COMSATS University Islamabad, Lahore Campus. 1.5 KM Defence Road off Raiwind Road, Lahore Pakistan**

   **COMSATS University Islamabad Lahore Campus, Lahore Pakistan**

   **Department of Computer Science**

3. **Gulsha Saleem**

   **Email: gulshansaleem26@gmail.com**

   **ORCID: 0000-003-2761-8399**

   **Address: Department of Computer Science, COMSATS University Islamabad, Lahore Campus. 1.5 KM Defence Road off Raiwind Road, Lahore Pakistan**

   **COMSATS University Islamabad Lahore Campus, Lahore Pakistan**

4. **Rana Hammad Raza**

   **Email: hammad@pnec.nust.edu.pk**

   **ORCID:000-0002-4883-7446**

   **Address: National University of Sciences and Technology (NUST) NUST-PNEC Habib Ibrahim Rehmatullah Road, Karachi**

5. **Muhammad Waqas Anwar**

**Email: waqasanwar@cuilahore.edu.pk**

**ORCID: 0000-0002-7822-8983**

**Address: Department of Computer Science, COMSATS University Islamabad, Lahore Campus. 1.5 KM Defence Road off Raiwind Road, Lahore Pakistan COMSATS University Islamabad Lahore Campus, Lahore Pakistan**

# Manuscript Title: Anomaly Recognition from Surveillance Videos using 3D Convolution Neural Network

*Abstract*— Anomalous activity recognition deals with identifying the patterns and events that vary from the normal stream. In a surveillance paradigm, these events range from abuse to fighting and road accidents to snatching, etc. Due to the sparse occurrence of anomalous events, anomalous activity recognition from surveillance videos is a challenging research task. The approaches reported can be generally categorized as handcrafted and deep learning-based. Most of the reported studies address binary classification i.e. anomaly detection from surveillance videos. But these reported approaches did not address other anomalous events e.g. abuse, fight, road accidents, shooting, stealing, vandalism, and robbery, etc. from surveillance videos. Therefore, this paper aims to provide an effective framework for the recognition of different real-world anomalies from videos. This study provides a simple, yet effective approach for learning spatiotemporal features using deep 3-dimensional convolutional networks (3D ConvNets) trained on the University of Central Florida (UCF) Crime video dataset. Firstly, the frame-level labels of the UCF Crime dataset are provided, and then to extract anomalous spatiotemporal features more efficiently a fine-tuned 3D ConvNets is proposed. Findings of the proposed study are twofold 1) There exist specific, detectable, and quantifiable features in UCF Crime video feed that associate with each other 2) Multiclass learning can improve generalizing competencies of the 3D ConvNets by effectively learning frame-level information of dataset and can be leveraged in terms of better results by applying spatial augmentation. The proposed study extracted 3D features by providing frame level information and spatial augmentation to a fine-tuned pre-trained model, namely 3DConvNets. Besides, the learned features are compact enough and the proposed approach outperforms significantly from state of art approaches in terms of accuracy on anomalous activity recognition having 82% AUC.

1. **INTRODUCTION**

Anomalous activity recognition from videos is one of the long-standing problems in computer vision and machine learning with wide-ranging applications in surveillance. Millions of surveillance cameras are being deployed in public and private places requiring intelligent video monitoring. While Video Content Analysis (VCA) has a very large collection of applications in a surveillance environment, one such focused area of research is anomalous activity recognition. Anomalous activity recognition deals with identifying the patterns and events that vary from the normal stream. Anomalies contain a huge range of activities that can go from abuse to fighting and road accidents to snatching such as shown in Figure 1. These events take place in different surroundings (such as public buildings and roads etc.) and during different times of the day and weather. Usually, anomalous activities occur infrequently as compared to normal activities. Therefore, developing an automated algorithm for video anomalous activity recognition is a pressing concern since manual handling of a high volume of video content is expensive and prone to error due to human limitations.

Recognition of anomalous events from unconstrained environments is a challenging problem. The challenges include huge Intra/inter-class variation where two or more classes have comparable characteristics, but fundamentally they are different. Moreover, insufficient annotated anomaly data and low resolution of surveillance videos are among the other challenges. Humans can perform recognition of these anomalies based on their common sense which is developed during years of learning. Machines, on the other hand, can identify such cases based on visual features that are learned using machine learning and deep learning. The machine learning algorithms can perform better if provided with unique features having better discriminatory power between different classes.

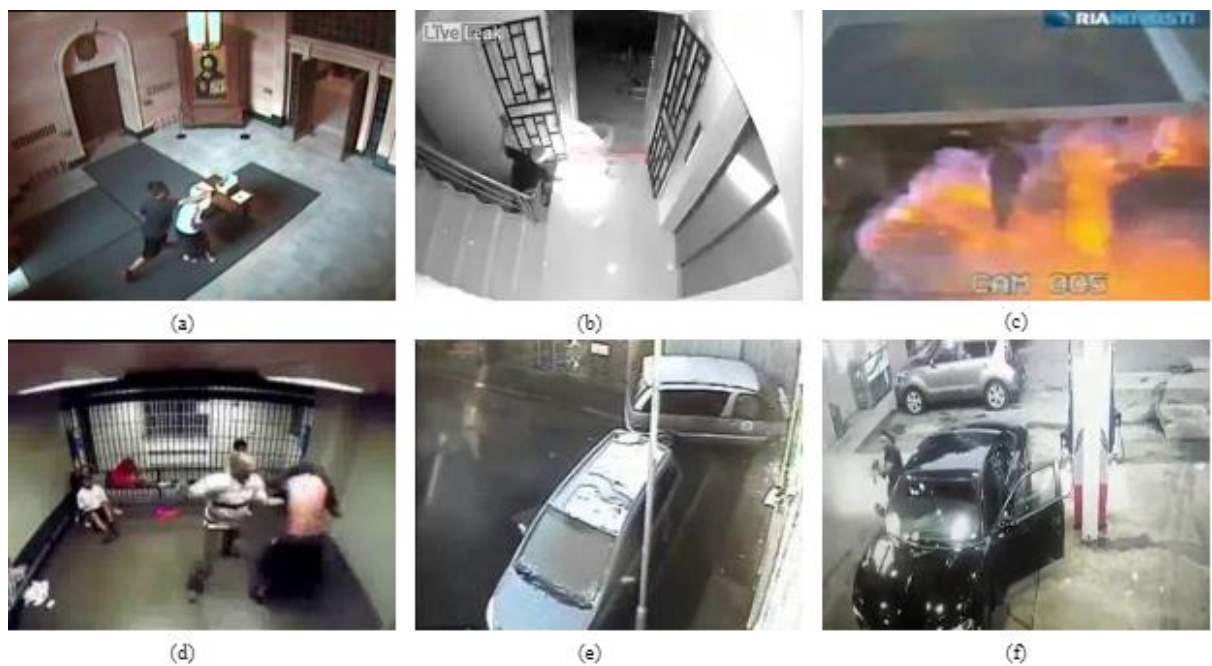

Figure 1: Sample anomalous frames from UCF Crime Dataset (a) Abuse (b) Arson (c) Explosion (d) Fight  (e) Road Accident (f) Shooting

For recognizing video anomalies, one category of popular methods in the literature with respect to deep learning is 3D ConvNets. Numerous recent studies have explored the dominance of 3D ConvNets based feature activation maps where human 2D/3D pose information is gained using depth algorithms [1]. The main idea of such architectures is to distinguish between different events, particularly by learning spatiotemporal features. Also in [2], the author factorized a 3D convolution kernel into a 2D spatial convolution kernel and 1D temporal convolution to learn depth features more efficiently. Most of the developed methods in the literature on activity recognition have focused on automated activities [3] where whenever any automated activity such as jogging, playing football, applying eye makeup, etc. occurs a deep learning-based architecture classifies it but this approach only focused on some generic activities. Also, most of the previously reported work identifies a particular anomalous activity, such as violence detection [4] [5], where substantial derivative equations are used as video descriptors to calculate local and convective acceleration (force) to detect violence from crowded places. Several studies have explored the detection of traffic accidents with different techniques including hybrid support vector machine with extended Kalman filter [6], stacked autoencoders [7], and Histogram of the gradient (HOG) [8]. However, such developed algorithms are not generalized enough to classify other anomalous activities, therefore they do not provide much use in practice. Besides, many researchers have used convolution networks along with spatiotemporal autoencoders [9] [10] for anomaly detection as a binary classification problem by coarsely classifying the surveillance videos as abnormal or normal. But their work did not provide possible solutions for differentiating between abnormal events, including explosion, abuse, fight, assault, stealing, robbery, and vandalism. Recently, a large scale CCTV dataset called UCF Crime dataset [11] was introduced that composed of 13 anomalous events captured from surveillance cameras. In [12] [13] multiclass classification is performed on this dataset and the reported accuracies are 23% and 31% respectively. In these proposed studies, experiments are conducted to address the problem of multiple anomalous events captured from surveillance cameras. To evaluate experiments, these studies have used the UCF Crime dataset. The dataset consists of a large spectrum that ranges from abuse to vandalism. To the best of our knowledge, none of the previous work except [12] [13] have worked for improving the multiclass classification accuracy on the UCF Crime dataset and their achieved accuracy was low for the recognition task.

Motivated by the aforementioned limitation of previous work, the goal of the proposed study is to learn spatiotemporal features from this dataset to provide predictions on these challenging videos. In the proposed study, features of video frames are analyzed for action recognition using a fine-tuned 3D ConvNets [3]. The learned features are efficient and robust enough to be competitive with existing approaches [12] [13] and the proposed study achieved state-of-art performance on the multi-class classification problem of the UCF Crime dataset. UCF Crime dataset consists of long untrimmed videos containing only video level annotation without any information about the temporal segment where the anomaly occurred. Previously reported study [12] provides annotation for only 140 videos and it treats the normal and anomaly-based videos as short bags but their proposed approach was not robust enough to address multiclass classification due to lack of annotation and spatial augmentation of the UCF Crime dataset. Annotating a video dataset is an exhaustive process. On the other hand, without frame-level and pixel-level information, deep network architecture cannot address accurate voting and prediction. The proposed study used a spatially augmented approach to the training set of this dataset by generating samples during the training process. Unlike previous approaches, the proposed system is trained on anomalous as well as normal events to provide better differentiation.

Contributions of the proposed study are following:

- ➢ The study provides frame-level annotations of the complete training set and applies spatial augmentation on the UCF Crime dataset.
- ➢ The proposed study uses spatial augmented features in a semi-supervised manner using fine-tuned 3D ConvNets [3] architecture.
- ➢ The experimental results on the dataset represent that as compared to the state-of-art anomalous activity recognition approaches the proposed study achieves higher performance in terms of accuracy.
- ➢ The proposed study provides a pilot study on different classes of UCF Crime dataset which will be useful for the modification and improvement of this dataset.

The remainder of this paper is arranged as follows. In Section II, the related work of anomalous activity recognition methods is summarized. Section III represents the proposed anomalous activity recognition method, where the spatially augmented and end-to-end fine-tuning of model is introduced. The used dataset and experimental results are given in Section IV and V, respectively. In the end, the discussion on the UCF Crime dataset and the conclusion is given in Section V1 and VII respectively.

## 2. RELATED WORK

A lot of work has been done for anomalous activity recognition from video streams. Anomalous activity recognition is a two-stage procedure that includes representative actions from the video stream and then the classification of learned action descriptors. We review anomaly detection/recognition-based methods mostly focusing on two groups. The first one is based on handcrafted approaches and the second one is deep learning-based approaches. In handcrafted based approaches spatiotemporal features of video dataset are extracted manually. Different well-known handcrafted algorithms such as the Histogram of Optical Flow(HOF), Hidden Markov Model (HMM) [14], the Histogram of Oriented Gradient (HOG), and optical flow are used for the extraction of features. After extraction of features from these models, extracted features are given as input to a classification model. A study [15] proposed a method for detection of an anomaly as the location-based approach, in comparison to the traditional object-based methods where features of an object are first identified then decisions for classification were made. They introduce an architecture that characterizes and model the object behavior at the pixel level based on motion information. To get motion information of objects, the background was subtracted. In the proposed method for motion properties of pixels, a Gaussian model was also fixed between spatiotemporal features extractor. The abnormal patterns were identified by using an HMM. Their proposed work contains information about direction such as the speed or size of the object. The motion of pixels that differ from other pixels was considered to be abnormal. They used the confusion matrix as an evaluation measure and achieved 91.25% results. But their method has two disadvantages such as, optical flow methods are computationally too complex for real-time applications without any special hardware and second is background subtraction mostly results in the subtraction of sensitive part of the image. Another study [16] proposed a framework that was used to identify multidimensional data of traffic for the analysis of different vehicle patterns and anomaly detection. This method was implemented using the dataset of traffic collected from different areas of cities. They proposed an architecture based on unsupervised learning hence no label for data was required. Their method was based on recognition of the anomaly pattern of the vehicle for real-time trajectory in videos in which they used Kalman Filter to avoid or overcome the occlusion problem. To learn spatial features an unsupervised technique Density-Based Spatial Clustering of Applications with Noise (DBSCAN) clustering was used that separates regular or irregular patterns from videos. But their architecture for detecting anomalies is not efficient is a term of computational cost at the testing time and their proposed model detects only one class such as traffic pattern. The author in [17] proposed a new feature descriptor called Histogram of Optical Flow and Magnitude Entropy (HOFME) to tackle the diverse anomalous scenarios. HOFME extracts spatiotemporal features from surveillance videos. These spatiotemporal features were based on the optical flow (OF) information to illustrate normal patterns in videos. For the classification of extracted features, they employed a simple nearest neighbor search to identify features as normal and abnormal. This study builds an approach of the 3D matrix for the information given by the optical flow field. Then each line and column represent orientation range and magnitude range respectively. The authors then used the entropy of the orientation flow to apprehend information on the appearance and density of anomalous regions [18].

The aforementioned handcrafted techniques are good for small datasets, but these approaches cannot handle large datasets efficiently. On the other hand, although many frameworks are based on handcrafted approaches, these approaches require over-engineering such as extraction of features, mid-level training as well as classification training. By considering these limitations of handcrafted approaches, this research work is conducted on a deep learning class called CNN that replicates all over engineering steps with a single neural network trained end to end. Deep learning is used in many applications and is evolving rapidly. Designing and altering network architectures to obtain desirable outcomes is a complex task. In many applications, the underlying network architectures remain mostly analogous i.e. without major customization. Though the plug and play process includes a combination of trial and error, varying primers, and data samples, etc. While the focus of this paper is on anomaly detection/recognition but to complete the discussion w.r.t deep learning, reference is being made to applications such as target detection [19], image generation [20], semantic segmentation [21], and text information hiding (steganography) [22], Transportation Scheduling System [23] where comparable deep learning architectures are used. 3D ConvNet displays outstanding performance for learning visual features by extracting the visual features and combining them, from a sequence of input images obtained from real-time video streams. These low-level features are then fed to the succeeding layers of 3D ConvNet to obtain high-level features. With the increasing popularity of 3DConvNet, studies in anomalous activity detection and recognition based on convolution neural networks are addressed below.

A study proposed [24] a 3D ConvNet based architecture and a bidirectional convolutional long short term memory model (ConvLSTM). This study presented deep learning-based gesture recognition architecture. The global and local features are encoded using 2D feature maps then, the higher features are extracted using 2DCNN. The spatiotemporal correlation of features was reserved throughout the feature extraction process. Again, a study [25] used 3D ConvNet to apprehend and store temporal information of crowd behavior. Their proposed 3D ConvNet consists of temporal as well as convolution and pooling layers. Various behavioral classes such as happy, sad, angry, excited, neutral, etc. are classified using support vector machine (SVM) algorithm. This study provided videos to 3D model as input that is broken into non-overlapping frames with the crowd (happy, sad, angry, excited, neutral, etc.) and the frames with no crowd known as Nothing. However, such studies [24] [25] address only short-range actions, as in UCF101 [26] dataset or HMDB [27] dataset. A study [28] proposed an anomalous activity recognition algorithm. For feature extraction of each cubic path in a video stream, the algorithm implements a cascaded network model. A group of Gaussian models is designed to classify anomalies. Simple cubic tracks use a straightforward neural network, and the continued cubic paths used a complicated and deep neural network for feature extraction. Again the same author in [29] proposed a study that applied a pre-trained fully convolution neural network for detection and localization of anomalies from videos. A fully connected network FCN was moved into an unsupervised FCN model ensuring the detection of anomalies from videos globally. It examined the cascaded detection for high performance in terms of speed and accuracy and addressed two main tasks i.e. anomalous feature illustration and cascaded outlier detection. Their method achieved AUC-ERR of 90.2% and 90.4% on the subway dataset [30] for both exit and entrance classes of the UMN dataset [31] respectively. Although these proposed, methods were good at achieving results on binary anomalous classification but not able to classify other anomalous classes. A study [32], proposed a deep network based framework for identification of the anomalous events in surveillance videos and used a single frame as an input to the CNN for feature learning. For binary classification, it used two trained CNN to capture the information from both spatial and temporal dimensions of videos. After the extraction of features, a final score of classification was assigned by fusing obtained scores from the two streams. The studies achieved 99.1% and 91% accuracy for the spatial and temporal stream of airport scenarios, respectively. Although, the study achieved remarkable results by using two streams of CNN, but the proposed model was only able to detect particular scenario based anomalies i.e. only airport environment based anomalous events. Moreover, authors at [3] proposed an unsupervised method that depicts the importance of a 3D ConvNets over 2DCNN. The reported work demonstrated that 3D architecture outperforms the traditional CNN and discussed the details of 3D ConvNets that have 3D kernels of size 3 x 3 x 3. These 3D kernels in every layer extract the finest features as compared with the 2D kernels and then these features were fed to pooling and fully connected layers. For classification purposes, the proposed study used a single linear classifier. The author used different action recognition-based datasets and achieved remarkable results. Although this study achieved remarkable results on different video based activity datasets, but the unsupervised approaches require improvement in the accuracy of behavior detection in different situations. Another model [33] named STF-Net was used for anomaly detection in surveillance videos using the UCSD dataset [34]. Spatiotemporal features were extracted using the STF-Net model that is identical in structure with 3D ConvNets. The first four layers of this model were used for the primary feature extraction of the video stream. Additionally, to enhance the variety of complex features(high-level) and improve the responsive field related to complex features for expressing the anomalous actions of minor and medium-sized levels, a structure of multi-scale was included in the STF-Net. After the extraction of the high spatiotemporal features of a video by using STF-Net, a mixed Gaussian model was trained on anomalous data. To detect whether the input video was anomalous, the distance between the extracted features was calculated using Mahalanobis based distance metric. Although the proposed model was good to cover the lack of temporal and spatial feature adaption, however, it was not able to classify other anomalous activities and only classified the crowded scenes as an anomaly.

Moreover, as the usage of pre-trained CNN based models is super incorporative and achieved solid results in the literature. Therefore, we reviewed brief recent work that has explored the advantage of using pre-trained models rather than building a CNN from scratch. A study [35] used a pre-trained 3DCNN model that produced spatiotemporal short-term features. After extraction of these consecutive time-varying features, LSTM was used to accumulate the dynamic behavior. Similarly, the author in [36] explored a technique to detect unusual events in surveillance videos using a pre-trained CNN. A pre-trained vgg16 based on a fully convolution neural network was used to learn spatial features for normal and abnormal behaviors. The author explored two approaches to detect anomalies. The first approach was homogenous, where a pre-trained model was utilized to fine-tuned CNN for each dataset. In the second approach (hybrid approach), a pre-trained model was utilized to fine-tuned CNN for one dataset which was further utilized to fine-tune another dataset. They evaluated their results on two benchmark datasets i.e. UMN dataset and USCD dataset for binary classification. The tube convolution neural network (T-CNN) [37] expands the recursive convolution neural network (R-CNN) [38] for the detection of objects from a video stream. It replicates the last layer with the tube of interest (TOI) pooling layer. Short snippets of videos are learned using this 3D deep model that is why this model was unable to capture the temporal depth of the dataset efficiently.  The author in [37] proposed workflow for the detection of anomalies from surveillance videos and used two-stream Deep Neural Networks (DNN) for the extraction of features. It used a pre-trained vgg16 model that comprised of 13 convolutional and three fully connected layers to classify an image extracted from video into 1000 image classes. The spatial features of videos were extracted using vgg16 and then the next procedure was to pass these extracted spatial features to the Long Short-Term Memory (LSTM) model to extract temporal features. The study used the UCF Crime dataset to calculate results against anomaly detection but only used four anomalous events for training. Therefore, the proposed work was not able to distinguish between other anomalous events. Another study [9] presents anomaly detection from videos in which a novel approach is used to present video data by a group of general features, that are deduced spontaneously from long video footage via a convolutional spatiotemporal autoencoder model. The proposed architecture comprises of two components 1) for spatial feature extraction, spatial autoencoder was used for each frame of video, and 2) for extracting temporal patterns of encoded spatial features, temporal encoder-decoder was used. The proposed spatial encoder and decoder comprises of two convolutional and deconvolutional layers respectively and the temporal encoder was a three-layer convolutional Long-Short-Term Memory (LSTM) model. The presented anomaly detection model was evaluated using UCSD dataset, and Subway dataset, but all of these datasets are environment-dependent and unable to generalize to other cases. Moreover, this approach needs an appropriate threshold to categorize reconstructed errors, and then it defines whether the video contains anomaly or not and cannot classify the type of anomaly. Sultani *et al.* [12] proposed a multiple instance learning approach (MIL) for anomaly detection and proposed a large scale crime based dataset called UCF Crime dataset. Labeling of frames is needed to detect anomalies from videos thereby making it a time-taking procedure. In the paper, authors have investigated the usage of a weakly labeled dataset, that has videos labeled as normal and anomalous (even if few frames have an anomaly and the remaining are normal frames) i.e. there is no frame level annotation available for this dataset. In this workflow, anomalous and normal videos were considered as positive and negative bags respectively and for feature extraction and classification process, a pre-trained 3DCNN model was employed. This model was specifically designed for the recognition of anomalous actions from the video stream, which was very incorporative for extracting both motion, and appearance-based features. The study achieved impressive results for binary classification but for anomalous activity recognition (multiclass classification), it did not perform well. Again a study [13] presented a Motion aware Tube Convolution Neural Network TCNN model for anomaly detection using the UCF Crime dataset. It combined the temporal framework into the Multiple-Instance Learning (MIL) ranking paradigm by using an attention block. The extracted attention features were used to differentiate between anomalous and normal segments of video streams. Furthermore, a deep neural network-based flow estimator PWCNet was used to calculate the optical flow between neighboring frames of a video stream. Binary classification, as well as multi-class classification, was done on the UCF Crime dataset and the achieved accuracy for multi-class classification was 31%.

After conducting an extensive literature review our study leads us to the conclusions that not much focus has been given to the classification of multiple anomalous events since the majority of the architectures are environment-dependent and fine-tuning of a pre-trained model is incorporative to achieve better results for anomalous activity recognition tasks. This motivates us to classify multiple anomalous events by comprehensively considering both the spatial and temporal context of videos. The proposed study makes a comparison with two approaches [12] [13] that did not achieve good results on the multi-classification of the UCF Crime dataset. The proposed work described in section 3 is not reliant on any particular environment and has the capability to classify certain types of anomalies. Different statistical experiments such as receiver operating characteristic (ROC), the area under the curve (AUC), and confusion matrix (CM) are conducted to explore the importance of the proposed work.

## 3. PROPOSED METHODOLOGY

The proposed method begins by dividing each video into a fixed number of frames. All the frames are first pre-processed and then 3D cubes are formed for model training. A combination of the fixed length of consecutive frames is given as input to the model. Both spatially augmented and original frames are passed to our fine-tuned model. The frames are not converted to grayscale so that channel information can be preserved. The proposed fine-tuned model consists of different numbers of convolution, pooling, batch normalization (BN), and fully connected layers (FC). The complete flow diagram of the proposed anomalous activity recognition approach is provided in Figure 2 where after preprocessing, 3D cubes are given to 3D ConvNets to extracts spatiotemporal features for each video clip, and a fully connected neural network is trained by exploiting a multiclass classifier SoftMax which calculates the probability between the maximum scored instances as shown in Figure 2 below. Both training and testing are semi-supervised i.e. only frame-level labels are present during training and evaluation. The output at the classification phase is given with the highest probability of predicted class. In this section, we first introduce the pre-processing phase, spatial augmentation technique, deep model, and its parameters for fine-tuning and then learning of the model.

### 3.1 PRE-PROCESSING

In the pre-processing phase, each video is first converted to its corresponding frames. These consecutive frames are then resized to 170×170 dimensions.

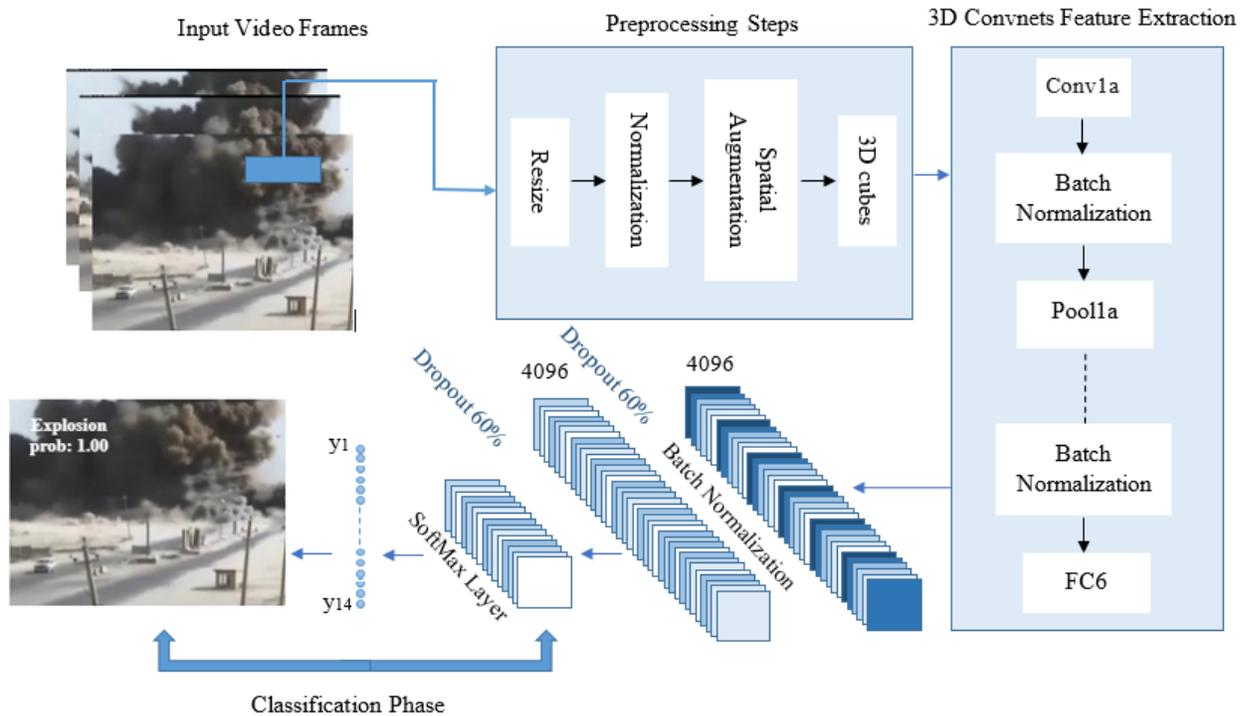

Figure 2: Flow diagram of the proposed anomalous activity recognition approach

Pixel values of all frames are then scaled between 0 to 1 to make sure that all frames are on the same scale. In the next step, every frame is then normalized, and spatial augmentation is applied to every frame of class as explained in section 3.2. After applying spatial augmentation, we make 3D cubes of each video into a fixed length to pass both spatial and temporal domains to our fine-tuned deep architecture.

### 3.2 SPATIAL AUGMENTATION

Deep learning needs a large amount of data for training purposes but due to limited dataset size, different methods are exploited to increase the number of examples. The procedure of simulating new dataset examples which preserve the accurate labels when limited labeled data exists is called data augmentation. The main goal of data augmentation is to make invariant predictions by covering different views of the same scene. Data augmentation for image classification typically depends on linear transformations in the spatial domain. For dataset based on time-series, data augmentation is more difficult because applying any transformation without domain knowledge may affect the temporal information of data. Recently, many studies have explored augmentation [38] [39] in data space or, as presented more recently, in feature space. Accurate augmentation for video-based datasets requires the availability of annotated data that contains information of each training sample. As we have performed frame-level annotation of each training instance of the UCF Crime dataset, so we applied the spatial augmentation technique on each training video.

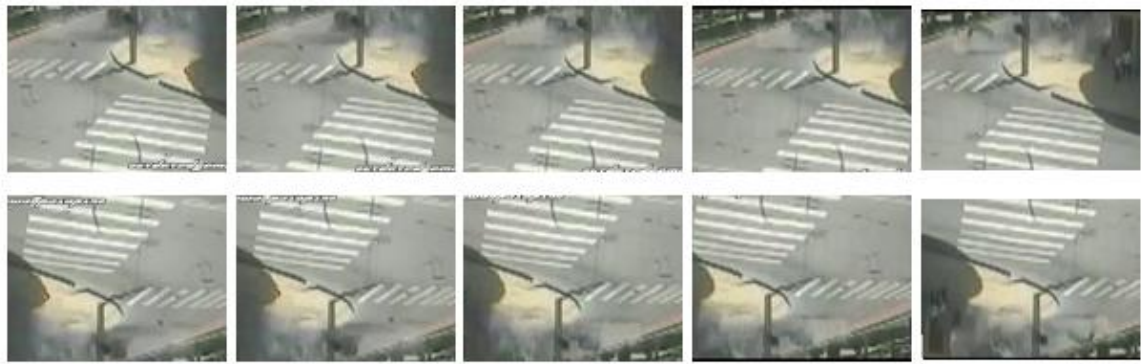

Figure 3: Original UCF Crime dataset frames (Top) and vertical spatial augmented frames (Bottom)

In the proposed study, augmentation on each video is performed in such a way that each video is first converted to a fixed number of frames $V = \{f_1, f_2, f_3, \ldots, f_n\}$ then each frame is augmented using horizontal and vertical flip techniques as shown in Figure 3. We have applied only these two video augmentation techniques, as augmenting a video dataset sometimes leads to the creation of noise in the dataset and affects the temporal information of the videos as well. Horizontal and vertical flipping is very successful [40] for image-based datasets so we applied these two techniques spatially to our video dataset so that temporal, as well as local information, may preserve.

**Flip Augmentation**: The horizontal and vertical flips are applied on extracted frames of each video, where each frame is first augmented using both techniques and to preserve temporal information accurately each frame is then converted to the corresponding video, as illustrated in Figure 4 that each frame is produced by spatially sampling with vertical and horizontal techniques and then converted to a video.

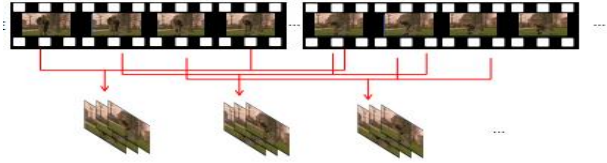

Figure 4: Illustration of our proposed spatial augmentation scheme.

To distinguish between various classes, particular spatio-temporal 3D filters are used to learn and illustrate various activity patterns. To extend the functionality of traditional CNN [41] for learning of spatiotemporal data, 3D convolutional kernels were proposed in which 3D kernels $K \in \mathbb{R}^{n_x \times n_y \times n_t}$ are convolved over a video cube $V \in \mathbb{R}^{m_x \times m_y \times m_t}$ that can be written as

$$F_{st} = V * K \qquad (1)$$

Where $*$ in Eq. 1 represents convolution and $F_{st}$ is the resultant spatiotemporal feature set. Usually, a learned kernel $K$ converts some low-level spatiotemporal action patterns such that the whole set of action patterns of each video cubes can be recreated from adequately many such convolving kernels. 3D ConvNets comprised of several pairs of the 3D convolution kernel and 3D pooling layers. These layers are then followed by several successive fully connected layers and in the end a softmax layer for multiclass classification to produce the desired output.

### 3.3 FINE TUNED 3DCONVNET MODEL

Fine-tuning is a conception of transfer learning where knowledge obtained during training in one kind of problem is used to train other related problems. In deep learning, training time while inheriting valued features can be saved by fine-tuning an existing model on a related problem with transfer learning techniques instead of developing a model from scratch.

The analysis on 3D kernels conducted in the above section motivates us to fine-tune the existing 3D ConvNets [3], in which a 3D ConvNets architecture was proposed that learned features in spatiotemporal domains of videos.

The reported architecture was trained on different video activity-based datasets and achieved remarkable results. For anomalous activity recognition on the UCF Crime dataset, we faced difficulties such as overfitting while using the same pre-trained model. Therefore, for optimal learning on the UCF Crime dataset, we applied transfer learning to existing 3D ConvNets architecture and classified 14 classes of the dataset with updated weights. For fine-tuning of the model three Batch Normalization layers were introduced at different stages of the model which proved to be useful for reducing overfitting and normalizing the output activations of the layers and the 3DConvNets after fine-tuning is shown in Table 1.

Batch Normalization (BN) is implemented to pre-trained models to gain the accurate speed, performance, and stability of architecture. During fine-tuning, the BN is achieved through normalizing steps by calculating the mean and variance of each activation layer across each mini batch. Preferably, the normalization is accompanied over the entire training set but it's impractical to use global information. Thus, BN is restrained to mini batch in the training set. For BN **B** denotes a mini-batch of size **m** for the entire training set then the empirical mean and variance for mini-batch is achieved as $\mu_B \leftarrow \frac{1}{m}\sum_{i=1}^{m} x_i$ and $\sigma_B^2 \leftarrow \frac{1}{m}\sum_{i=1}^{m}(x_i - \mu_B)^2$ for the layers of the model with d-dimensional input, $x = (x^{(1)}, \ldots, x^{(d)})$ where $\mu_B$ and $\sigma_B^2$ are mean and variance respectively. By applying BN, the iterative weights update starts with a group of randomly initialized weights. Before the starting of the training stage, each convolutional layer of 3D ConvNets weights is initialized by values arbitrarily sampled from a normal distribution with a zero mean and small standard deviation. Additionally, fine-tuning was done using the Stochastic Gradient Descent (SGD) algorithm and an initial learning rate of 0.001 and a momentum of 0.09.

Table 1: Details of fine-tuned 3D ConvNets architecture used in our experiments

| Layer | Input | Kernel | Output |
|---|---|---|---|
| Input | 16×170×170×3 | N/A | 16×170×170×3 |
| conv1 | 16×170×170×64 | 3×3 | 16×170×170×64 |
| batchNormalization_1 | 16×170×170×64 | -- | 16×170×170×64 |
| pool1 | 16×170×170×64 | 1×2 | 16×85×85×64 |
| conv2 | 16×85×85×64 | 3×3 | 16×85×85×128 |
| pool2 | 8×85×85×128 | 2×2 | 8×43×43×128 |
| conv3a | 8×43×43×128 | 3×3 | 8×43×43×256 |
| conv3b | 8×43×43×256 | 3×3 | 8×43×43×256 |
| pool3 | 8×43×43×256 | 2×2 | 4×22×22×256 |
| conv4a | 4×22×22×256 | 3×3 | 4×22×22×512 |
| conv4b | 4×22×22×512 | 3×3 | 4×22×22×512 |
| pool4 | 4×22×22×512 | 2×2 | 2×11×11×512 |
| conv5a | 2×11×11×512 | 3×3 | 2×11×11×512 |
| conv5b | 2×11×11×512 | 3×3 | 2×11×11×512 |
| pool5 | 2×13×13×512 | 2×2 | 1×6×6×512 |
| batchNormalization_2 | 1×6×6×512 | -- | 1×6×6×512 |
| fc6 | (None,18432) | 1×1 | (None,4096) |
| batchNormalization_3 | (None,4096) | -- | (None,4096) |
| fc7 | (None,4096) | 1×1 | (None,4096) |
| fc9 | (None,4096) | 1×1 | (None,14) |

## 4 DATASET DESCRIPTION

Different datasets exist for anomalous activity recognition from videos and most of them represent different automated actions e.g. in the UMN dataset [31] five different types of dramatic videos are present in which people running in various directions are classified as an anomaly. In UCSD Ped1 and Ped2 datasets [34], the videos only reflect simple events and do not address the real anomalies e.g. walking on walkways and any non-pedestrian entities such as wheelchair and biker in the walkways are considered as an anomaly.

Due to the aforementioned limitations of datasets, we used a recently presented anomaly-based large scale dataset called the UCF Crime dataset [11]. This dataset consists of a total of 1900 long untrimmed videos that contain 13 real-world anomalies namely, abuse, arrest, arson, assault, accident, burglary, explosion, fighting, robbery, shooting, stealing, shoplifting, and vandalism. Out of 1900 videos, 950 of them contain normal videos and the remaining 810 are anomalous videos. Most of the videos of this dataset consist of multiple anomalies categories. For example, robbery with fighting, arrest with shooting and road accident, and assault with fighting. All the dataset videos consist of realistic real-world surveillance applications.

This dataset provides approximately 129 hours of videos having a resolution of 320 x 240. These videos are not normalized in lengths. We selected this dataset for multi-class classification because these anomalies have a significant effect on public safety. But this dataset has two major limitations, 1) Video level labels i.e. we only know that each video contains anomaly, but the exact segment of anomaly is not known and 2) anomalous classes of this dataset have huge inter-class variations [42] that become the reason for overfitting during training. For multiclass classification, both comparison-based approaches [12] [13] have used the same training and testing split where the training set contains 38 videos of each class, and the testing set contains 12 videos of each class. We have used the same testing set but our training set contains both original and spatially augmented videos.

Table 2: Statistics of spatially augmented UCF Crime Dataset for our training

|  | Statistics of Spatially Augmented UCF Crime dataset | |
|---|---|---|
|  | Anomalous | Normal |
| Number of videos | 1040 | 80 |
| Total length (sec) | 86.34 | 7.627 |
| Min/ Max length (sec) | 0.6/1165.0 | 7.62/3600 |
| Average length (sec) | 56.60 | 184.90 |

**Frame-level annotations for UCF Crime dataset:** To the best of our knowledge, none of the previous [12] [13] anomaly detection approaches provide frame-level annotations for these unusual events in their training set. To overcome the paucity of labeled data, we improve the training set of the recently proposed UCF-Crime dataset with frame-level annotation. We benefit from the fact that normal videos do not need any further annotation therefore, we annotate all anomaly videos with frame-level annotation so that a semi-supervised approach can be used for training and evaluation of 3DConvnets. The new training set of videos contains frame level information of anomaly segments.

**Training and testing set:** Each class of this dataset comprises only 38 videos in its training set which are not sufficient for accurate model learning. To overcome this limitation, we augment each video spatially as discussed in the augmentation section. The statistics of the spatially augmented dataset are given above in Table 2 where maximum and minimum video clip lengths are presented for training set only. We do not make any change in the test set of the UCF Crime dataset thus it consists of 168 videos. The training and testing sets of this dataset comprise of all 13 anomalous classes and normal class videos. Each anomalous class is discriminative enough due to frame level annotation. Moreover, the frame distribution of the training set used in our experiment after spatial augmentation and length are shown in Figure 5. Also, the frame distribution of the UCF Crime training set used in comparison based approaches [12] [13] is shown in Figure 6.

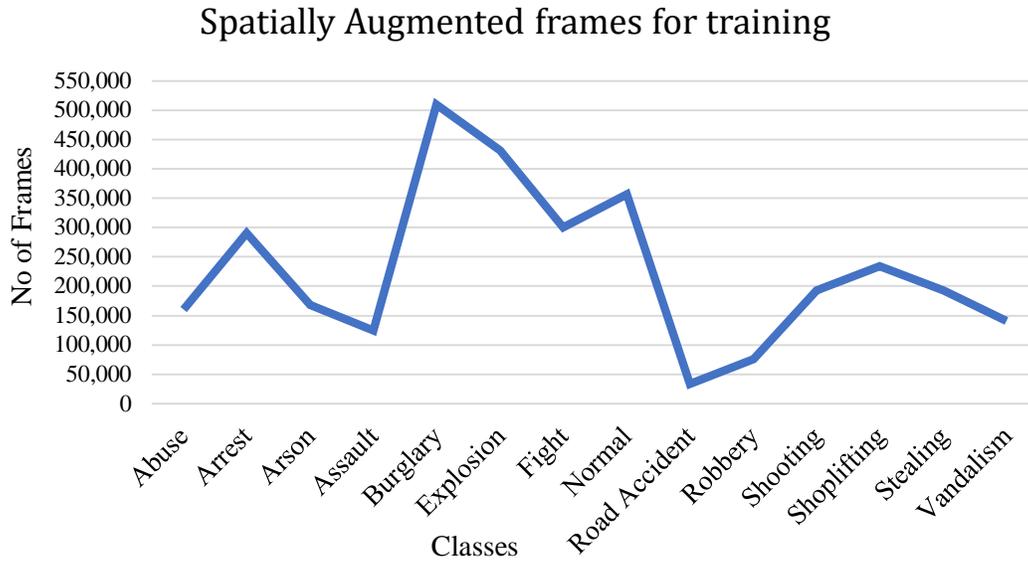
Figure 5: No of frames used in experiments

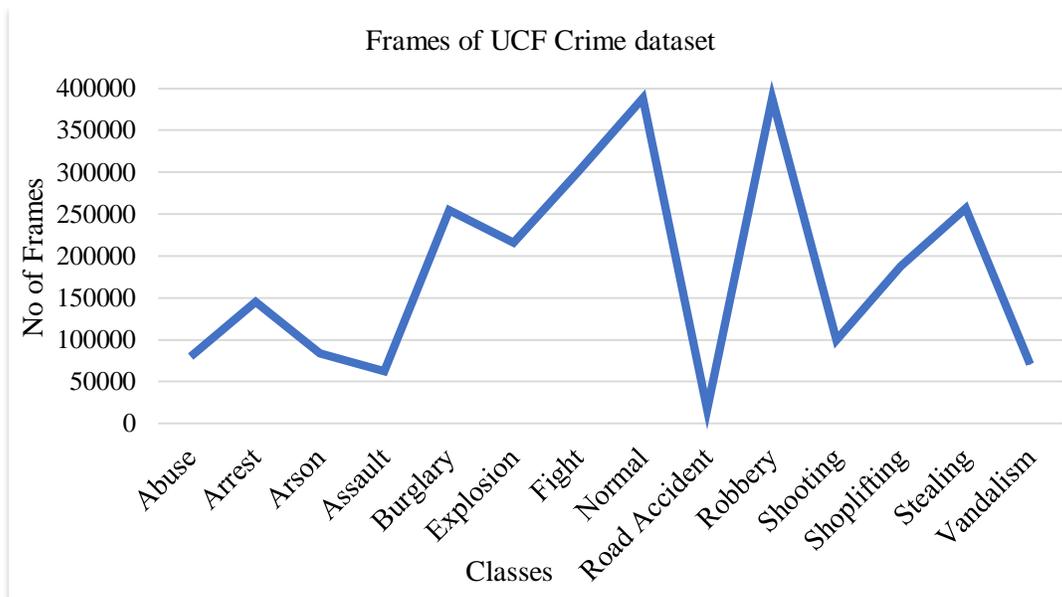
Figure 6: Distribution of frames in the training set of UCF Crime

## 5 RESULTS AND EXPERIMENTS

In this section, we tested the proposed fine-tuned 3DConvNets model on the UCF Crime dataset. Results for multiclass classification are conducted on this dataset. The proposed solution is also compared with the recent state of the art methods. In addition to that, different statistical metrics are used to evaluate the advantage of our work. Moreover, at the end of this section, a detailed discussion on the limitations of the UCF Crime dataset is carried out. We also resized frames to 200 × 200 but did not find any improvement in results that is why for less consumption of computational resources, 170×170 dimensions size of frames were used.

## 5.1 IMPLEMENTATION DETAILS:

**Feature Extraction:** For the extraction of features, we first performed all pre-processing steps as explained in section 3.1. Pre-processing is done to make the dataset ready for model input. Each video frame is resized to 170×170 with the 30fps fixed frame rate. We used 16 frames of cubes (motion information) to compute features and normalized each cube. To extract spatiotemporal features, these 16-frames cubes are then passed to our proposed fine-tuned 3D ConvNets model. The first layer of the model takes 4D input such as 16×170×170×3 where 16 and 3 represent the total number of frames passed at a time and RGB channel, respectively. This input is fed to the first convolution layer that extracts spatiotemporal features by convolving 3D kernel over the input of 16-frames cube. Batch Normalization layer is applied after the 1$^{st}$ layer of convolution to compute features that prove useful for reducing overfitting and normalizing the output activations of a layer. We also tried different layers by increasing model size but unfortunately, they did not observe any improvement in recognition accuracy. The features are extracted by convolving different convolution layers followed by pooling and batch normalization layers. Details of model layers and their parameters are given in Table 3 where dashes show the absence of parameters in the relevant layer.

After extraction of spatiotemporal features, these extracted features are fed to a three-layered fully connected neural network, and a 60% dropout is applied between these FC layers. These fully connected (FC) layers transformed the extracted features into a 4096-dimensional feature vector. At the classification stage, the softmax activation function is used that contains output units, which resulted in anomalous activity recognition.

Table 3: 3DConvNets parameters description

| Parameters | Layers | |
|---|---|---|
| | 3D Convolution | 3D Max pooling |
| Size | $3 \times 3 \times 3$ | $2 \times 2 \times 2$ |
| Subsampling | $1 \times 1 \times 1$ | - |
| Activation Function | Relu | - |
| Border mode | Same | Valid |
| Stride | - | $1 \times 2 \times 2$ |

**Proposed Model Implementation:** This proposed network is implemented using Keras implementation of 3D ConvNets. All the model layers are implemented using Keras built-in libraries of deep learning. After setting all required layers of the model as explained in section 3.4, we implemented different hyperparameters for the training of the 3D model. After trying several parameters of 3D ConvNets, the performance and efficiency of the model to learn robust features is attained by using mini batches of 32 and cubes to 16 frames which also supports being less computationally expensive for the less consumption of resources. Stochastic Gradient Descent (SGD) optimizer results best for our model. In SGD optimizer, the learning rate was set to .0001 that updates after every 3 epochs if no change was found during the learning of the model. Another parameter of SGD called the momentum was set to 0.09. The tuning of all these parameters was used to compute loss and accuracy for each class during model learning. The analysis of our model is conducted using Ge-Force GTX 1080ti (11GB) connected with 352-bit memory on a desktop PC with 2.10-GHz Intel Core i5-3310M and 16GB RAM.

**Inference Phase:** At the inference stage, we take a clip and then preprocess it by resizing it into a specific resolution at which our model is trained. We calculate the input tensors length by dividing each video clip by the total sample size. Then we input the clip data to the trained model and loop the output tensors where each tensor has 14 values. These 14 values represent scores for the 14 classes. In the end, we get the index for the top value and calculate percentages and put these values into the respective 16 frames. In Table 4, we evaluate our trained model on a real-world video and have shown some intermediate frames of video with its predicted class. The video length is 49 seconds having a total of 810 frames. Our trained model predicts the highest probability class on every 16 cubes of video such as each predicted class is shown with its frame occurrence.

Table 4: Model prediction on intermediate frames of real-world video

| Intermediate Frames from Real word Video | Prediction/ Frame No | Actual Class |
|---|---|---|
| 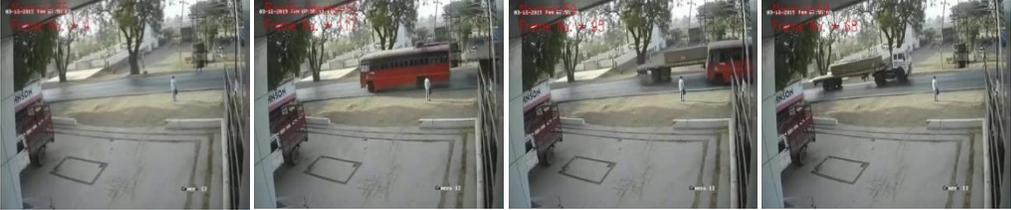 | Normal/ Frame 04-68 | Normal |
| 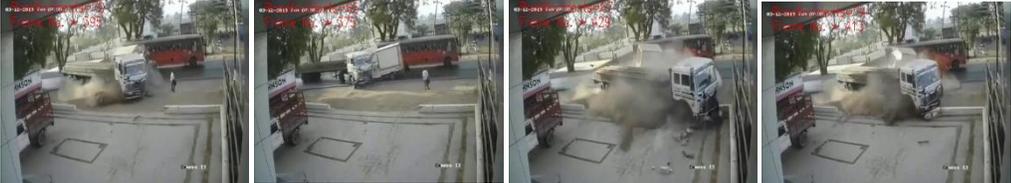 | Road Accident/ Frame 395-413 | Road Accident |
| 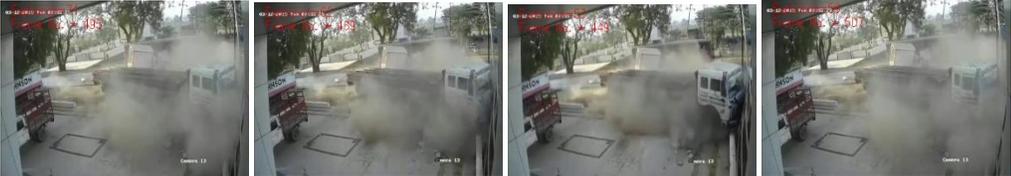 | Explosion/ Frame 495/507 | Explosion |

**Evaluation Metric:** By following published approaches on anomalous activity recognition, the proposed study uses the four evaluation matrices. The accuracy, confusion matrix, receiver operating characteristic (ROC) curve for each activity, and corresponding area under the curve (AUC) are used to evaluate the performance of our approach. Micro average and macro average AUC are calculated to aggregate the contribution of all classes for evaluation of multiclass classification more accurately. Also, to show the mean performance of the 3DConvNets classifier, more statistical measures such as precision, recall, and F-measure (trade-off between precision and recall) are used.

## 5.2 RESULTS ON UCF CRIME DATASET

For a multiclass classification problem, the Area Under the ROC curve has gained much success. This statistical metric is desirable because during the training process misclassification cost and class distribution are not known. By considering the importance of the AUC-ROC curve in machine learning, we evaluate the performance of our model using this metrics. The AUC-ROC curve of fine-tuned 3DConvNets on the UCF Crime dataset is shown in Figure 7 where the true positive rate (on the y-axis) is plotted against false positive rate (on the x-axis). An excellent model has AUC near 1 which means it has good separability between all classes and AUC near 0.5 means the model has poor separability between classes [43]. AUC evaluated on our proposed model is near 1 for abuse, arson, explosion, fight, normal, road accident, robbery, shoplifting, and vandalism classes. For shooting, stealing, and assault, AUC is near 0.7 which shows our model performance is good for these classes too. But for arrest and burglary, AUC is near 0.5 which shows the misclassification of the classifier due to the intra-class similarity issues of the UCF Crime dataset.

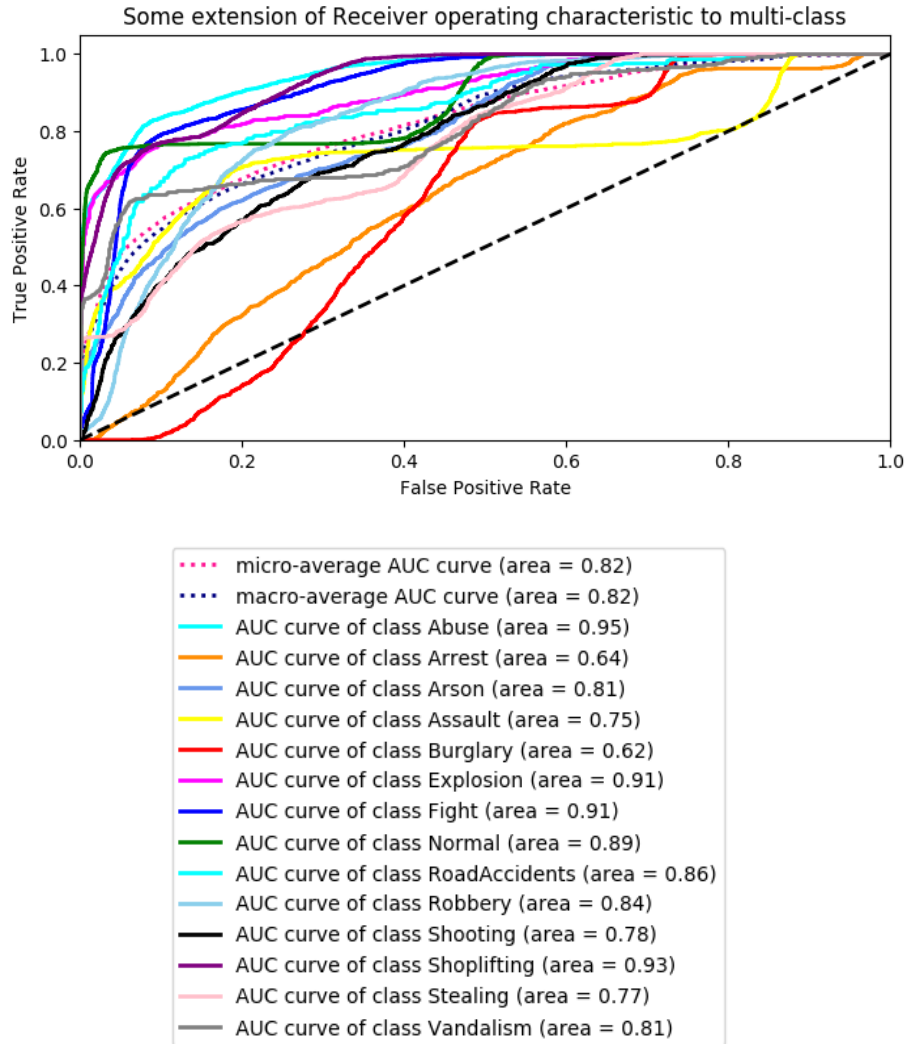

Figure 7: AUC-ROC curve obtained for the multiclass classification of UCF Crime dataset using fine-tuned 3DConvNets

Micro-AUC and Macro-AUC are also used to determine the overall performance of the proposed model across the dataset.

Another well-known statistical metric, the confusion matrix is used to evaluate proposed model performance. The confusion matrix of the 3DConvNets on the UCF Crime dataset is shown in Table 5, where the main diagonal represents the correctly predicted responses and most of the anomaly classes, as well as normal class, are almost predicted well. Also, an average anomalous activity recognition accuracy of fine-tuned 3DConvNets on the UCF Crime dataset is 45%.

Table 5: Resultant Confusion Matrix of our approach showing the per-class accuracy on UCF Crime dataset

| Classes | Abuse | Arrest | Arson | Assault | Burglary | Explosion | Fight | Normal | Road Accidents | Robbery | Shooting | Shoplifting | Stealing | Vandalism |
|---|---|---|---|---|---|---|---|---|---|---|---|---|---|---|
| Abuse | **0.61** | 0 | 0.072 | 0.023 | 0.01 | 0.057 | 0 | 0.021 | 0 | 0.1 | 0 | 0 | 0 | 0 |
| Arrest | 0.048 | **0** | 0.066 | 0.047 | 0.034 | 0 | **0.18** | 0.0102 | **0.17** | 0.01 | **0.13** | 0 | 0 | 0 |
| Arson | 0.023 | 0 | **0.30** | 0 | 0.014 | 0 | 0.01 | 0.11 | 0 | 0.12 | 0.21 | 0.048 | 0 | 0.091 |
| Assault | 0.024 | 0 | 0.031 | **0.26** | 0 | 0 | **0.28** | 0.026 | 0 | 0 | 0.01 | 0 | 0 | 0 |
| Burglary | 0 | 0.13 | 0 | 0 | **0.053** | 0 | 0.015 | 0 | 0.1 | 0.23 | 0.02 | 0.12 | 0.21 | 0.069 |
| Explosion | 0 | 0 | 0.058 | 0 | 0 | **0.60** | 0 | 0 | 0.12 | 0 | 0.11 | 0 | 0 | 0.004 |
| Fight | 0.012 | 0 | 0 | 0.015 | 0.17 | 0 | **0.66** | 0 | 0 | 0 | 0 | 0.0063 | 0 | 0 |
| Normal | 0.0133 | 0 | 0 | 0 | 0 | 0 | 0 | **0.78** | 0 | 0 | 0 | 0.023 | 0 | 0.23 |
| Road Accidents | 0.062 | 0 | 0 | 0.016 | 0.016 | 0.016 | 0.041 | 0.033 | **0.60** | 0.078 | 0 | 0 | 0.016 | 0 |
| Robbery | 0 | 0 | 0.011 | 0 | 0 | 0.011 | 0.094 | 0 | 0.009 | **0.50** | 0 | 0.29 | 0 | 0.0084 |
| Shooting | 0 | 0 | 0.0046 | 0.0046 | 0 | 0.027 | 0.003 | 0.1 | 0.06 | 0 | **0.27** | 0 | 0 | 0.0066 |
| Shoplifting | 0 | 0 | 0.096 | 0 | 0 | 0 | 0.008 | 0 | 0 | 0.11 | 0 | **0.80** | 0.0031 | 0 |
| Stealing | 0 | 0 | 0.036 | 0.009 | 0 | 0 | 0.019 | 0 | 0 | 0.24 | 0 | 0.21 | **0.30** | 0 |
| Vandalism | 0 | 0 | 0.025 | 0 | 0 | 0 | 0.012 | 0 | 0.23 | 0 | 0 | 0 | 0.0406 | **0.60** |
| Average Accuracy (%) | | | | | | | | | | | | | | **45%** |

To determine the mean performance of proposed 3DConvNets, we conduct the additional statistical experiment by using precision, recall, f-measure shown in Table 6. These statistical experiments also performed well which indicates that the model classifier returned most of the relevant classes correctly. All evaluation measures demonstrated that only abuse, assault, arson, fight, normal, explosion, road accidents, robbery, and shoplifting classes are well distinguishable from the proposed model. Statistical metrics did not perform well on other classes such as arrest, burglary, and shooting due to the intra-class similarities. These classes contain other anomalies such as arrest videos from the test set of the UCF Crime dataset contains shooting, fight, and road accident anomalies. A detailed discussion about the dataset is provided in section 6.

Table 6: Statistical metrics obtained for UCF Crime dataset using 3DConvNets

| Classes | Precision | Recall | F-measure |
|---|---|---|---|
| Abuse | 0.81 | 0.62 | 0.73 |
| Arrest | 0.00 | 0.00 | 0.00 |
| Arson | 0.45 | 0.35 | 0.45 |
| Assault | 0.61 | 0.40 | 0.45 |
| Burglary | 0.09 | 0.05 | 0.07 |
| Explosion | 0.71 | 0.55 | 0.62 |
| Fight | 0.70 | 0.66 | 0.70 |
| Road Accident | 0.40 | 0.54 | 0.45 |
| Robbery | 0.35 | 0.45 | 0.45 |
| Shooting | 0.24 | 0.28 | 0.26 |
| Shoplifting | 0.51 | 0.74 | 0.45 |
| Stealing | 0.37 | 0.45 | 0.45 |
| Vandalism | 0.40 | 0.55 | 0.55 |
| Normal event | 0.67 | 0.75 | 0.71 |
| **Average** | **0.45** | **0.45** | **0.45** |

Table 7: Multiclass Accuracy Comparison of UCF Crime dataset between different approaches

| Method | Accuracy |
|---|---|
| C3D [12] | 23% |
| TCNN [12] | 29% |
| Motion+TCNN [13] | 31% |
| **Proposed method** | **45%** |

### 5.3 COMPARISON WITH THE STATE-OF-THE-ART

The results obtained from the fine-tuned 3DConvNets are compared quantitatively with two states of the art methods to measure the effectiveness of the proposed anomaly approach. Authors at [12] proposed an anomalous activity recognition method using C3D and TCNN to classify only anomalous classes to recognize anomalies. They did not use normal videos during their experiments. Moreover, the achieved results for anomalous activity recognition can be improved. Similarly, a motion based TCNN [13] was used for the improvement of the anomalous activity recognition task. The study proposed a model for anomalous activity recognition by using the MIL approach as of study [12] due to the limitations of frame-level annotation for the UCF Crime dataset. This study only used anomalous classes for classification during training and at the testing stage due to which their trained model was not able to distinguish normal events from abnormal. A well-known multiclass classifier called SoftMax is used as a baseline method. Particularly, we use both anomalous and normal videos for training so that our model can learn better visual features for both patterns. 3D ConvNets features are calculated for each clip, and then we trained a multiclass classifier with a linear kernel. At the testing stage, the classifier gives the highest probability of each video batch to perform classification. Based on the comparison, Table 7 demonstrates that the proposed method obtained the best results in terms of accuracy (%) on the UCF Crime dataset.

## 6  DISCUSSION

In this section, we will investigate our results for anomalous activity recognition on the UCF Crime dataset collected from evaluation matrices to provide a pilot study on this dataset. As can be seen in Table 5 and Figure 7, there is 0% accuracy and 0.64% AUC for class arrest which shows there exists no class discrimination for arrest class. In terms of visualization too, we observe the limitations of the UCF Crime dataset for multiclass classification. Apart from evaluation matrices, we provide some of the observations and visual inspection of different videos of the dataset here:

*Arrest018_x264.mp4*: No clear activity of arrest happens. The video starts with an accident after which police cars arrive, but no clear activity of arrest happens in the whole video. *Arrest016_x264.mp4:* In this video from frame number 88 to 131 there was a police officer who points a gun at a person having a flag in his hand. From frame 132 to 208 fighting occurs and the remaining video contains just shooting incidents. No arrest activity occurs. From *Arrest017_x264.mp4 to Arrest020_x264.mp4* no act of arrest occurs but these videos contain the act of anomaly where police officers are tracking a culprit. Therefore, our confusion matrix fails to classify arrest class and most of the frames are classified as Fight, Road Accidents, and shooting as shown in Table 5 and also to explain the findings qualitatively Table 8 is showing some examples from the UCF Crime dataset that have intraclass similarity with other activities. The actual anomaly class column represents the UCF Crime definition for each class and both intraclass similarity columns are representing feature similarities with other classes.

Table 8: Qualitative explanation of intraclass similarity of UCF Crime dataset **[11]**

| File Info | Actual Anomaly Class | Intra Class Similarity 01 | Intra Class Similarity 02 |
|---|---|---|---|
| Arrest016_x264.mp4 | This class comprises of videos that represent police arresting people. | Fighting<br>Frame 132: 208<br>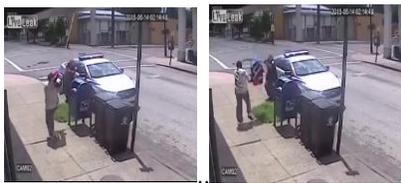<br>People attacking each other (Fighting definition of UCF Crime dataset) | Shooting<br>Frame 208: 290<br>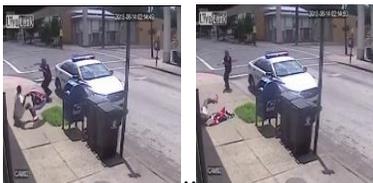<br>The action of shooting somebody with a gun (Shooting definition of UCF Crime dataset) |
| Arrest017_x264.mp4 | | Road Accident<br>Frame 229:290<br>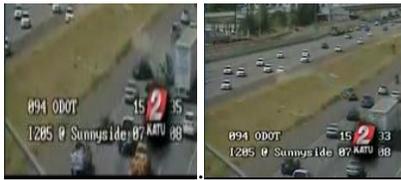 | -- |

| | | Road accidents including vehicles, walkers, or cyclists (Road Accident definition of UCF Crime dataset). | |
|---|---|---|---|
| Assault008_x264.mp4 | This class comprises of videos in which people that are assaulted never fight back | Fighting<br>Frame 10:250<br>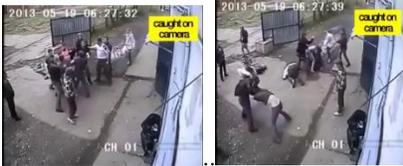<br>People are attacking each other (Fight definition of UCF Crime dataset) | -- |
| Fight006_x264.mp4 | This class comprises of videos in which human attacks each other. | Assault<br>Frame 151: 215<br>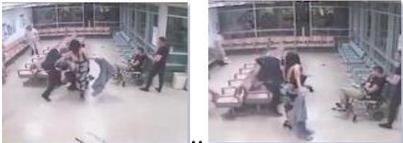<br>The assaulted person does not fight back (Assault definition of UCF Crime dataset) | -- |
| Burglary010_x264.mp4 | This class comprises videos in which people entered a shop/building with intention of theft. | Stealing<br>Frame 130:349<br>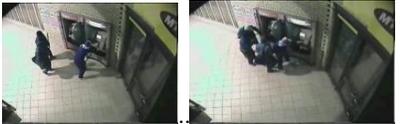<br>People are taking goods/money illegally (Stealing definition of UCF Crime dataset) | Robbery<br>Frame 10251:10559<br>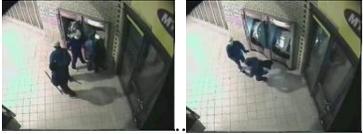<br>Thieves taking money illegally (Robbery definition of UCF Crime dataset) |
| Burglary009_x264.mp4 | | Stealing<br>Frames 114: 180<br>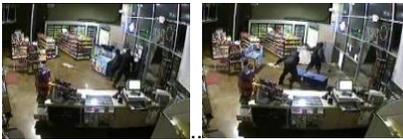<br>People are taking goods illegally (Stealing definition of UCF Crime dataset) | -- |

We found evidence in favor of our results by visually inspecting other videos of the dataset e.g. class Fight and Assault. In this dataset, most of the assault videos contain the activity of fighting in it. For example, *Assault008_x264.mp4* and *Fight006_x264.mp4* contain the same event scenarios. *Assault008_x264.mp4* starts with fighting between a group of people. After 5 sec of the video, the assaulted persons start fighting with the opponents. The same scenario for *Fight006_x264.mp4* where the video starts with a person coming towards a man with the intention of fighting and after 2 seconds in the video, the fight starts between them as shown in Table 8. Almost 20 to 25 videos of Assault class have features similar to fight class.

Our visual inspection and results show that the UCF Crime dataset needs improvement for future work in multiclass classification due to huge intra-class variation between different classes of this dataset. Striking a balance between accuracy over the UCF-Crime without overfitting is one of the major tasks for anomalous activity recognition.

## 7  CONCLUSION

This research work mainly focuses on the recognition of multiple anomalies from surveillance videos to eliminate much human intervention. This study proposes an automated deep learning-based approach for real-world anomalous activity recognition. This work is conducted because not much work has been done so far on the recognition of multiple anomalies and mostly researchers address only binary classification i.e. either a video is normal or containing abnormality. Our extensive literature review also shows that why deep learning-based approaches have superiority over handcrafted based approaches for the extraction of features from videos. The proposed study provides a fine-tuned, pre-trained 3D ConvNets architecture that outperforms on the previously reported approaches. This 3D model is used to efficiently extract both spatiotemporal features from surveillance videos. The study also addresses the importance of the existence of frame-level labeling for better learning of spatiotemporal features in a semi-supervised manner. Furthermore, this work has demonstrated the significance of spatial augmentation to gain better results when training a deep architecture. The proposed methodology is applied to the large-scale UCF Crime dataset. The experiments conducted on this dataset demonstrate that the fine-tuned 3DConvNets outperforms the existing state-of-art anomalous activity recognition approaches in terms of accuracy. The proposed work also provides a pilot study about different classes of the UCF Crime dataset and discussed its limitations for the anomalous activity recognition task that will be helpful for future work on this dataset.

## 8  ACKNOWLEDGMENTS

We acknowledge partial support from the National Center of Big Data and Cloud Computing (NCBC) and HEC of Pakistan for conducting this research.

# 9 REFERENCES

[1] T. Huynh-The, H. Hua-Cam and D. S. Kim, "Encoding Pose Features To Images With Data Augmentation For 3D Action Recognition," *IEEE Transactions on Industrial Informatics.*, vol. 16, no. 5, pp. 3100 - 3111, April 2019.

[2] L. Sun, K. Jia, D. Y. Yeung and B. E. Shi, "Human action recognition using factorized spatio-temporal convolutional networks," in *Proceedings of the IEEE international conference on computer vision*, ICCV, 2015, pp. 4597--4605.

[3] D. Tran, L. Bourdev, R. Fergus, L. Torresani and M. Paluri, "Learning spatiotemporal features with 3d convolutional networks," in *Proceedings of the IEEE international conference on computer vision*, ICCV, 2015, pp. 4489--4497.

[4] S. Mohammadi, H. Kiani, A. Perina and V. Murino, "Violence detection in crowded scenes using substantial derivative," in *2015 12th IEEE International Conference on Advanced Video and Signal Based Surveillance (AVSS)*, IEEE, 2015, pp. 1-6.

[5] T. Zhang, Z. Yang, W. Jia, B. Yang, J. Yang and X. He, "A new method for violence detection in surveillance scenes," *Multimedia Tools and Applications,* vol. 75, no. 12, pp. 7327-7349, 2016.

[6] V. M. Vishnu, M. Rajalakshmi and R. Nedunchezhian, "Intelligent traffic video surveillance and accident detection system with dynamic traffic signal control," *Cluster Computing,* vol. 21, no. 1, pp. 135--147, 2018.

[7] D. Sigh and C. K. Mohan, "Deep spatio-temporal representation for detection of road accidents using stacked autoencoder," *IEEE Transactions on Intelligent Transportation Systems,* vol. 20, no. 3, pp. 879--887, 2018.

[8] A. P. Shah, J. B. Lamare, T. Nguyen-Anh and A. Hauptmann, "CADP: A novel dataset for CCTV traffic camera based accident analysis," in *2018 15th IEEE International Conference on Advanced Video and Signal Based Surveillance (AVSS)*, IEEE, 2018, pp. 1-9.

[9] Y. S. Chong and Y. H. Tay, "Abnormal event detection in videos using spatiotemporal autoencoder," in *In Advances in Neural Networks - ISNN 2017 14th International Symposium*, Sapporo, Hakodate, and Muroran, Springer, 2017, pp. 189--196.

[10] M. Sabokrou, M. Fayyaz, M. Fathy , Z. Moayed and R. Klette , "Deep-anomaly: Fully convolutional neural network for fast anomaly detection in crowded scenes," *Computer Vision and Image Understanding,* vol. 172, pp. 88--97, 2018.

[11] University of Central Florida, "Real-world Anomaly Detection in Surveillance Videos," CVCR, 2011. [Online]. Available: https://www.crcv.ucf.edu/projects/real-world/. [Accessed 20 April 2020].

[12] W. Sultani, C. Chen and M. Shah, "Real-world anomaly detection in surveillance videos," in *Proceedings of the IEEE Conference on Computer Vision and Pattern Recognition*, 2018, pp. 6479--6488.

[13] Y. Zhu and S. Newsam, "Motion-Aware Feature for Improved Video Anomaly Detection," in *British Machine Vision Conference*, BMVC, 2019.

[14] R. Colque, C. Caetano, M. Andrade and W. Schwartz, "Histograms of optical flow orientation and magnitude and entropy to detect anomalous events in videos," *IEEE Transactions on Circuits and Systems for Video Technology,* vol. 27, no. 3, pp. 673--682, 2017.

[15] V. Mahadevan, W. Li, V. Bhalodia and N. Vasconcelos, "Anomaly detection in crowded scenes," in *2010 IEEE Computer Society Conference on Computer Vision and Pattern Recognition*, IEEE, 2010, pp. 1975--1981.

[16] M. Farooq, N. Khan and M. Ali, "Unsupervised video surveillance for anomaly detection of street traffic," *International Journal of Advanced Computer Science and Applications (IJACSA),* vol. 12, no. 8, pp. 270-275, 2017.

[17] R. Colque, C. Caetano, M. de Andrade and W. R. Schwartz, "Histograms of optical flow orientation and magnitude and entropy to detect anomalous events in videos," *IEEE Transactions on Circuits and Systems for Video Technology,* vol. 27, no. 3, pp. 673--682, 2016.

[18] A. A. Sodemann, M. P. Ross and B. J. Borghetti, "A review of anomaly detection in automated surveillance," *IEEE Transactions on Systems, Man, and Cybernetics, Part C (Applications and Reviews),* vol. 42, no. 6, pp. 1257--1272, 2012.

[19] Y. Tian, A. Dehghan and M. Shah, "On detection, data association and segmentation for multi-target tracking," *IEEE Transaction on patren analysis and machine inteligence,* vol. 41, no. 9, pp. 2146-2160, 2018.

[20] W. Cai and W. Zhango, "PiiGAN: Generative adversial networks for pluralistic image inpainting," *IEEE Access Remote sensing image recognition,* no. 8, pp. 48451-48463, 2010.

[21] H. You, S. Tian and L. Yu, "Pixel-level remote sensing image recognition based on bidirectional word vectors," *IEEE Transactions on Geoscience and Remote Sensing,* vol. 58, no. 2, pp. 1281-1293, 2019.

[22] Z.-L. Yang, X.-Q. Guo, Z.-M. Chen, Y.-F. Huang and Y.-J. Zhang, "RNN-stega: Linguistic stenography based on recurrent neural networks," *IEEE Transaction on Information Forensics and Security,* vol. 14, no. 5, pp. 1280-1295, 2018.25

[23] S. Zhang, C. Lu, S. Jiang, L. Shan and N. N. Xiong, "An Unmanned Intelligent Transportation Scheduling System for Open-Pit Mine Vehicles Based on 5G and Big Data," *IEEE Access,* vol. 8, pp. 135524--135539, 2020.

[24] L. Z. Zhang, S. Guangming , S. Peiyi , A. S. Juan and M. Bennamoun, "Learning Spatiotemporal Features Using 3DCNN and Convolutional LSTM for Gesture Recognition," in *Proceedings of the IEEE International Conference on Computer Vision (ICCV) Workshops*, 2017.

[25] E. Varghese and S. M. Thampi, "A deep learning approach to predict crowd behavior based on emotion," in *International Conference on Smart Multimedia*, Springer, 2018, pp. 296--307.

[26] G. Sigurdsson, O. Russakovsky and A. Gupta, "What actions are needed for understanding human actions in videos?," in *Proceedings of the IEEE international conference on computer vision*, 2017, pp. 2137--2146.

[27] S. C. Yu, S. Yun, C. Songzhi and L. S. Guorong , "Stratified pooling based deep convolutional neural networks for human action recognition," *Multimedia Tools and Applications,* vol. 76, no. 11, pp. 13367--13382, 2017.

[28] M. Sabokrou, M. Fayyaz, M. Fathy and R. Klette, "Cascading 3d deep neural networks for fast anomaly detection and localization in crowded scenes," *IEEE Transactions on Image Processing,* vol. 26, no. 4, pp. 1992--2004, 2017.

[29] M. Sabokrou, M. Fayyaz, M. Fathy, Z. Moayed and R. Klette, "Fully convolutional neural network for fast anomaly detection in crowded scene," *Computer Vision and Image Understanding,* vol. 172, pp. 88--97, 2018.

[30] Z. Andrei and . W. Richard , "Anomalous Behavior Data Set," Department of Computer Science and Engineering and Centre for Vision Research York University, Toronto, ON, Canada, [Online]. Available: http://vision.eecs.yorku.ca/research/anomalous-behaviour-data/. [Accessed 27 September 2020].

[31] University of Central Florida, "Abnormal Crowd Behavior Detection using Social Force Model," CVCR, 2011. [Online]. Available: https://www.crcv.ucf.edu/projects/Abnormal_Crowd/. [Accessed 20 April 2020].

[32] A. Jamadandi, S. Kotturshettar and U. Mudenagudi, "Two Stream Convolutional Neural Networks for Anomaly Detection in Surveillance Videos," in *Smart Computing Paradigms: New Progresses and Challenges*, Springer, 2020, pp. 41--48.

[33] Z. Li, Y. Li and Z. Gao, "Spatiotemporal Representation Learning for Video Anomaly Detection," *IEEE Access,* vol. 8, pp. 25531--25542, 2020.

[34] SVCL, "UCSD Anomaly Detection Dataset," Svcl, 2013. [Online]. Available: http://www.svcl.ucsd.edu/projects/anomaly/dataset.html. [Accessed 20 April 2020].

[35] B. Kim and J. Lee, "A deep-learning based model for emotional evaluation of video clips," *International Journal of Fuzzy Logic and Intelligent Systems,* vol. 18, no. 4, pp. 245--253, 2018.

[36] S. Bansod and A. Nandedhak, "Transfer learning for video anomaly detection," *Journal of Intelligent & Fuzzy Systems,* vol. 36, no. 3, pp. 1967-1975, 2019.

[37] U. Koppikar, C. Sujatha, P. Patil and U. Mudenagudi, "Real-World Anomaly Detection Using Deep Learning," in *International Conference on Intelligent Computing and Communication}*, Springer, 2019, pp. 333--342.

[38] T. T. Um, F. M. Pfister, D. E. Pichler, L. M. Satoshi , S. F. Hirche and K. D. Urban , "Data augmentation of wearable sensor data for Parkinson's disease monitoring using convolutional neural networks," in *Proceedings of the 19th ACM International Conference on Multimodal Interaction*, 2017, pp. 216--220.

[39] X. Cui, V. Geol and B. Kingsbury, "Data Augmentation for Deep Neural Network Acoustic Modeling,," *IEEE/ACM Transactions on Audio, Speech, and Language Processing,* vol. 23, no. 9, pp. 1469--1477, 2015.

[40] H. Lou, C. Xiong, W. Fang, P. E. Love, B. Zhang and X. Ouyang, "Convolutional neural networks: Computer vision-based workforce activity assessment in construction," *Automation in Construction,* vol. 94, pp. 282--289, 2018.

[41] Y. LeCun, L. Bottou, Y. Bengio and P. Haffner, "Gradient-based learning applied to document recognition," *Proceedings of the IEEE,* vol. 86, no. 11, pp. 2278--2324, 1998.

[42] M. R. Vilamala, L. Hiley, Y. P. Hicks and C. F. Alun, "A Pilot Study on Detecting Violence in Videos Fusing Proxy Models," *vilamala2019pilot,* 2019.

[43] S. Narkhede, "Understanding AUC-ROC Curve," *Towards Data Science,* vol. 26, pp. 220-227, 2018.

[44] I. Goodfellow, Y. Bengio and A. Courville, Deep learning, MIT press, 2016.

[45] H. Gao, B. Cheng, J. Wang , K. Li, J. Zhao and D. Li, "ObjeObject Classification Using CNN-Based Fusion of Vision and LIDAR in Autonomous Vehicle Environment," *IEEE Transactions on Industrial Informatics,* vol. 14, no. 9, pp. 4224--4231, 2018.

[46] R. Hou, C. Chen and M. Shah, "Tube convolutional neural network (T-CNN) for action detection in videos," in *Proceedings of the IEEE international conference on computer vision*, 2017, pp. 5822--5831.

[47] R. Girshick, J. Donahue, T. Darrell and J. Malik, "Rich feature hierarchies for accurate object detection and semantic segmentation," in *Proceedings of the IEEE conference on computer vision and pattern recognition*, 2014, pp. 580--587.